\documentclass{bmvc2k}
\usepackage{amsfonts}

\usepackage{multirow}
\usepackage{booktabs}
\usepackage{graphicx}
\usepackage{tabularx}
\usepackage{caption}
\usepackage{floatrow}

\newcommand{\overbar}[1]{\mkern 1.5mu\overline{\mkern-1.5mu#1\mkern-1.5mu}\mkern 1.5mu}

\newfloatcommand{capbtabbox}{table}[][\FBwidth]

\title{DwNet: Dense warp-based network for pose-guided human video generation}

\addauthor{Polina Zablotskaia}{pzablots@cs.ubc.ca}{12}
\addauthor{Aliaksandr Siarohin}{aliaksandr.siarohin@unitn.it}{3}
\addauthor{Bo Zhao}{bzhao03@cs.ubc.ca}{12}
\addauthor{Leonid Sigal}{lsigal@cs.ubc.ca}{12}

\addinstitution{
 Department of Computer Science\\
 University of British Columbia\\
 Vancouver, BC, Canada \\ 
}
\addinstitution{
 Vector Institute for Artificial Intelligence \\
 Toronto, ON, Canada
}
\addinstitution{
 DISI\\
 University of Trento\\
 Trento, Italy
}

\runninghead{Zablotskaia, Siarohin, Zhao \& Sigal}{DwNet: Dense warp-based network}

\def\eg{\emph{e.g}\bmvaOneDot}

\def\etal{\emph{et al}\bmvaOneDot}

\begin{document}

\maketitle

\begin{abstract}
Generation of realistic high-resolution videos of human subjects is a challenging and important task in computer vision. In this paper, we focus on human motion transfer -- generation of a video depicting a particular subject, observed in a single image, performing a series of motions exemplified by an auxiliary (driving) video. Our GAN-based architecture, DwNet, leverages dense intermediate pose-guided representation and refinement process to warp the required subject appearance, in the form of the texture, from a source image into a desired pose. Temporal consistency is maintained by further conditioning the decoding process within a GAN on the previously generated frame. In this way a video is generated in an iterative and recurrent fashion. We illustrate the efficacy of our approach by showing state-of-the-art quantitative and qualitative performance on two benchmark datasets: TaiChi and Fashion Modeling. The latter is collected by us and will be made publicly available to the community. 

\end{abstract}


\section{Introduction}
\label{sec:intro}

Generative models, both conditional and un-conditional, have been at the core of computer vision field from its inception. In recent years, approaches such as GANs \cite{goodfellow2014generative} and VAEs \cite{kingma2013auto} have achieved impressive results in a variety of image-based generative tasks. The progress on the video side, on the other hand, has been much more timid. Of particular challenge is generation of videos containing high-resolution moving human subjects. In addition to the need to ensure that each frame is realistic and video is overall temporally coherent, additional challenge is contending with coherent appearance and motion realism of a human subject itself. Notably, visual artifacts exhibited on human subjects tend to be most glaring for observers (an effect partially termed "uncanny valley" in computer graphics). 

\begin{figure}[t]
\centering
\includegraphics[width=1.0\textwidth]{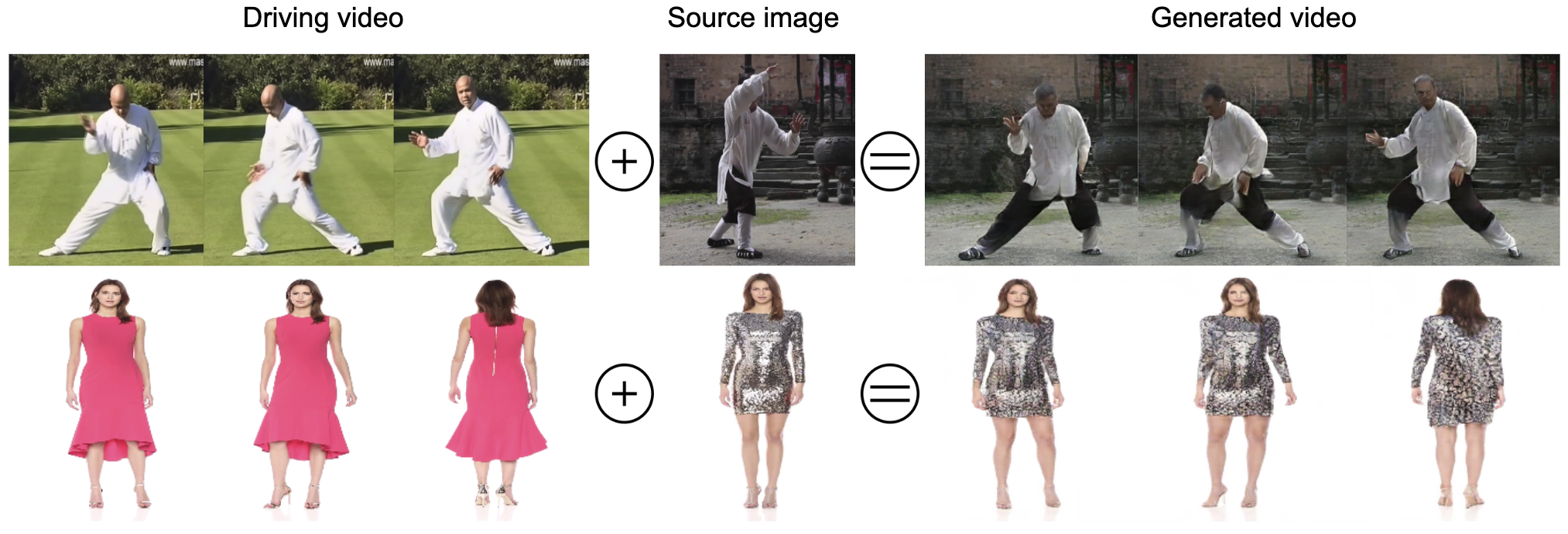}
\caption{{\bf Human Motion Transfer.} Our pose-guided approach allows generation of a video for the source subject, depicted in an image, that follows the motion in the driving video.}
\label{fig:teaser}
\vspace{-0.07in}
\end{figure}

In this paper, we address a problem of human motion transfer. Mainly, given a single image depicting a (source) human subject, we propose a method to generate a high-resolution video of this subject, conditioned on the (driving) motion expressed in an auxiliary video. The task is illustrated in Figure~\ref{fig:teaser}. 
Similar to recent methods that focus on pose-guided image generation \cite{ma2017pose, siarohin2018deformable, balakrishnansynthesizing, ma2018disentangled, esser2018variational, dong2018soft, zanfir2018human,Neverova_2018_ECCV,grigorev2019coordinate,siarohin2019animating}, we leverage an intermediate pose-centric representation of the subject. 
However, unlike those methods that tend to focus on sparse keypoint \cite{ma2017pose, siarohin2018deformable, balakrishnansynthesizing, ma2018disentangled, dong2018soft} or skeleton \cite{esser2018variational} representations, or intermediate dense optical flow obtained from those impoverished sources \cite{siarohin2019animating}, we utilize a more detailed dense intermediate representation \cite{Guler2018DensePose} and texture transfer approach to define a fine-grained warping from the (source) human subject image to the target poses. This texture warping allows us to more explicitly preserve the appearance of the subject. 
Further, we focus on temporal consistency which ensures that the transfer is not done independently for each generated frame, but is rather sequentially conditioned on previously generated frames. 
We also note that unlike \cite{chan2018everybody, wang2018video}, we rely only on a {\em single} (source) image of the subject and not a video, making the problem that much more challenging. 

\vspace{0.1in}
\noindent
{\bf Contributions:} Our contributions are multiple fold. First, we propose a dense warp-based architecture designed to account for, and correct, dense pose errors produced by our intermediate human representation. Second, we formulate our generative framework as a conditional model, where each frame is generated conditioned not only on the source image and the target pose, but also on previous frame generated by the model. This enables our framework to produce a much more temporally coherent output. Third, we illustrate the efficacy of our approach by showing improved performance with respect to recent state-of-the-art methods. Finally, we collect and make available a new high-resolution dataset of fashion videos.

\section{Related work}
\label{sec:related}





\noindent \textbf{Image Generation.}
Image generation has become an increasingly popular task in the recent years. The goal 
is to generate realistic images, mimicking samples from true visual data distribution. 
Variational Autoencoders (VAEs)~\cite{kingma2013auto} and Generative  Adversarial  Networks (GANs)~\cite{goodfellow2014generative} are powerful tools for image generation that have shown promising results. In the case of the unconstrained 
image generation, resulting images are synthesized from random noise vectors. However, this paradigm can be extended to the {\em conditional} image generation~\cite{pix2pix2016, pix2pixHD}, where apart from the noise vector the network input includes conditional information, which can be, for example, a style image \cite{pix2pix2016,johnson2016perceptual}, a descriptive sentence \cite{hong2018textim} or an object layout \cite{zhao2019layoutim} designating aspects of desired image output. Multi-view synthesis~\cite{zhou2016view, park2017transformation, sun2018multi} is one of the largest topics in the conditional generation and it is the the one that mostly related to our proposal. The task of multi-view synthesis is to generate unseen view given one or more known views.


\vspace{0.1in}
\noindent \textbf{Pose guided image generation.}
In the pioneering work of Ma~\etal~\cite{ma2017pose} pose guided image generation using GANs has been proposed. Ma~\etal~\cite{ma2017pose} suggest to model human poses as a set of keypoints and use standard image-to-image translation networks (\eg, UNET~\cite{ronneberger2015u}). Later, it has been found~\cite{balakrishnansynthesizing, siarohin2018deformable} that for UNET-based architectures it is difficult to process inputs that are not spatially aligned. In case of pose-guided models, keypoints of target are not spatially aligned with the source image. As a consequence, \cite{balakrishnansynthesizing, siarohin2018deformable} propose new generator architectures that try to first spatially align these two inputs and then generate target images using  image-to-image translation paradigms.  Neverova~\etal~\cite{Neverova_2018_ECCV} suggest to exploit  SMPL~\cite{SMPL:2015} representation of a person, which they estimate using DensePose~\cite{Guler2018DensePose}, in order to improve pose-guided generation. Compared to keypoint representations, DensePose~\cite{Guler2018DensePose} results provide much more information about the human pose, thus using it as a condition allows much better generation results. Grigorev~\etal~\cite{grigorev2019coordinate} propose coordinate based inpainting to recover missing parts in the DensePose~\cite{Guler2018DensePose} estimation. Coordinate based inpainting explicitly predicts from where to copy the missing texture, while regular inpainting predicts the RGB values of the missing texture itself. Contrary to Grigorev~\etal~\cite{grigorev2019coordinate}, our work can perform both standard inpainting and coordinate based inpainting. 

\vspace{0.1in}
\noindent \textbf{Video Generation.}
The field of video generation is much less explored, compared to the image generation. Initial works adopt 3D convolutions and recurrent architectures for video generation~\cite{vondrick2016generating, saito2017temporal, tulyakov2017mocogan}. Later works start to focus on conditional video generation. The most well studied task, in conditional video generation, is future frame prediction~\cite{srivastava2015unsupervised,oh2015action,finn2016unsupervised, babaeizadeh2017stochastic}. Recent works 
exploit intermediate representation, in the form of learned keypoints, for future frame prediction~\cite{villegas2017learning, zhao2018learning, wang2018every}. However, the most realistic video results are obtained by conditioning video generation on another video. This task is often called video-to-video translation. Two recent works~\cite{chan2018everybody, wang2018video} suggest pose-guided video generation, which is a special case of video-to-video translation. The main drawback of these models is that they need to train a separate network for each person. In contrast, we suggest to generate a video based only on a {\em single} image of a person. Recently, this task was addressed by Siarohin~\etal~\cite{siarohin2019animating}, but they try to learn a representation of a subject in an unsupervised manner which leads to sub-optimal results. Conversely, in our work, we exploit and refine the richer structure and representation from DensePose~\cite{Guler2018DensePose} as an intermediate guide for video generation.

\section{Method}
\label{sec:method}
\label{sec:over}

\begin{figure}[h]
\begin{tabular}{cc}
\bmvaHangBox{\includegraphics[width=5.6cm]{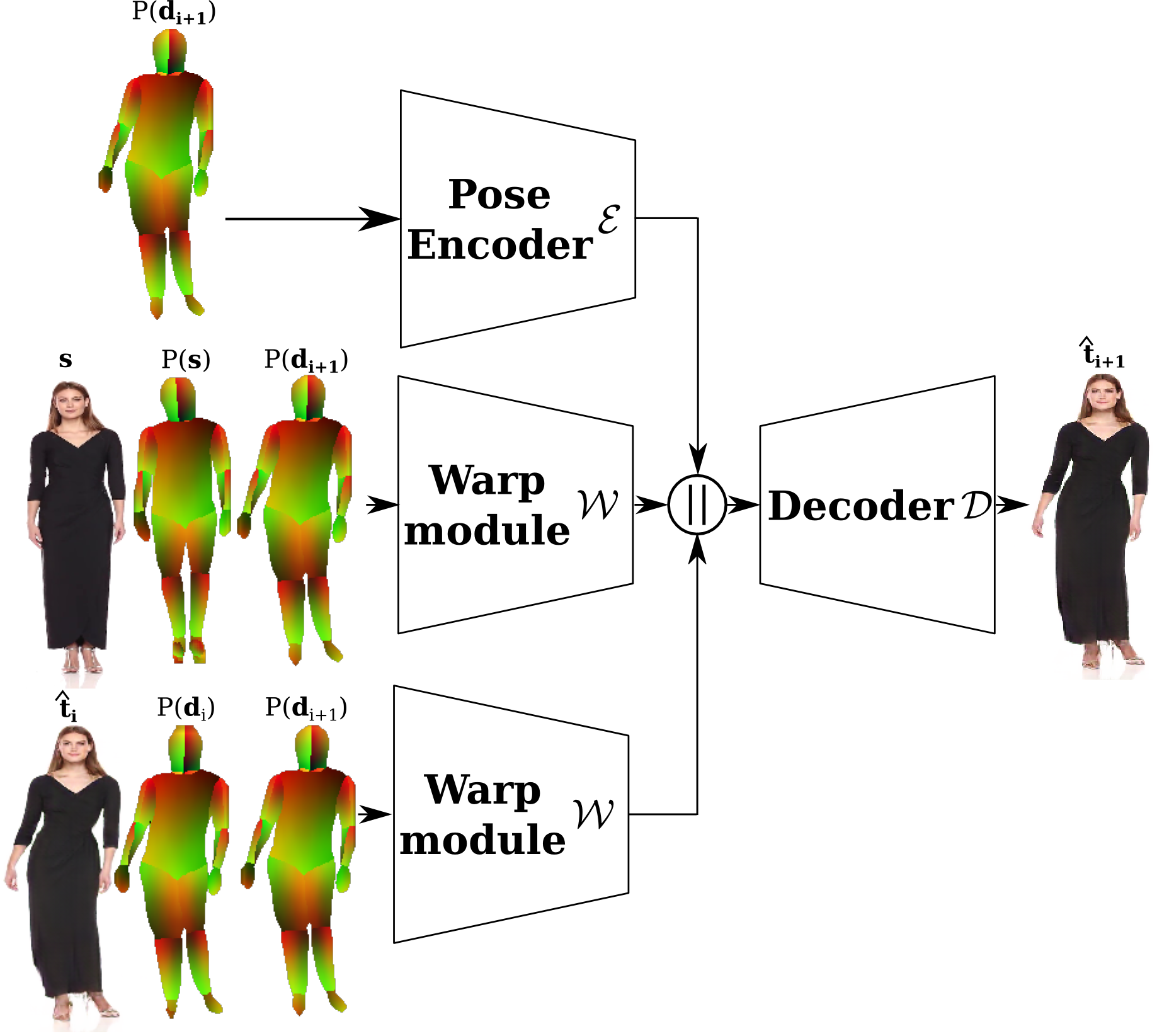}} &
\bmvaHangBox{\includegraphics[width=5.6cm]{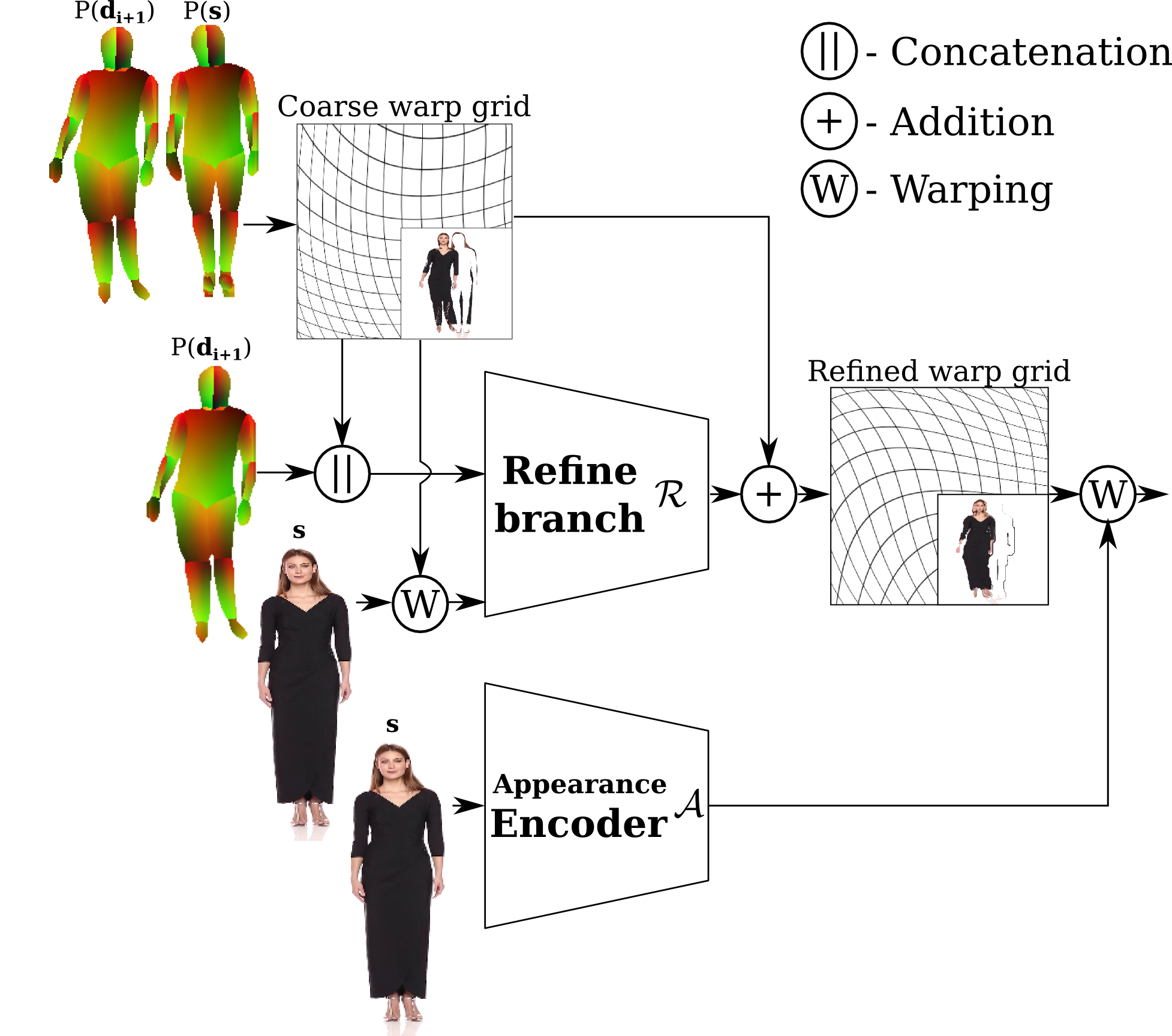}}\\
(a)&(b)
\end{tabular}
\caption{{\bf Proposed Architecture.} Overview of the entire model (a). The model consists of three main parts: Pose Encoder ($\mathcal{E}$), Warp Module ($\mathcal{W}$) and Decoder ($\mathcal{D}$). The encoder $\mathcal{E}$ takes as input the next driving video pose $P(\mathbf{d}_{i+1})$ and produces its feature representation. On the other hand source image $\mathbf{s}$, source image pose $P(\mathbf{s})$ and the next frame's pose $P(\mathbf{d}_{i+1})$ are used by $\mathcal{W}$ to produce deformed feature representation of the source image $\mathbf{s}$. Similarly, $\mathcal{W}$ produces a deformed feature representation of previously generated frame $\mathbf{\hat{t}}_i$. Decoder $\mathcal{D}$ combines these representations and generates a new frame $\mathbf{\hat{t}}_{i+1}$. Detailed overview of the warp module $\mathcal{W}$ is illustrated in (b). In the beginning, we compute a coarse warp grid from $\mathcal{D}_{i+1}$ and $P(\mathbf{s})$. The coarse warp grid is used to deform the source image $\mathbf{s}$, this deformed image along with driving pose $P(\mathbf{d}_{i+1})$ and coarse warp coordinates is used by Refine branch $\mathcal{R}$ to predict correction to coarse warp grid. We employ this refined estimate to deform feature representation of the source image $\mathbf{s}$. Note, deformation of $\mathbf{\hat{t}}_i$ takes similar form.}
\label{fig:overview}
\end{figure}


The objective of this work is to generate a video, consisting of $N$ frames $\left[\mathbf{t}_1, \dots, \mathbf{t}_N \right]$, containing a person from a {\em source} image $\mathbf{s}$,  conditioned on a sequence of driving video frames of another person $\left[\mathbf{d}_0, \dots, \mathbf{d}_N\right ]$, such that the person from the source image replicates motions of the person from a driving video. Our model is based on standard image-to-image~\cite{pix2pix2016} translation framework. However, the standard convolutional networks are not well suited for the task where the condition and the result are not well aligned. Hence, as advocated in the recent works \cite{balakrishnansynthesizing,siarohin2018deformable}, the key to a precise human subject video generation lies in leveraging of motion from the estimated poses. Moreover, perceptual quality of the video is highly dependent on the temporal consistency between nearby frames. We design our model having these goals and intuitions in mind (see Figure~\ref{fig:overview}(a)). 

First, differently from standard pose-guided image generation frameworks~\cite{ma2017pose, siarohin2018deformable, balakrishnansynthesizing}, in order to produce temporary consistent videos we add a markovian assumptions to our model. If we generate each frame of the video independently, the result can be temporary inconsistent and have a lot of flickering artifacts. To this end, we condition generation of each frame $\mathbf{t}_i$ on a previously generated frame $\mathbf{t}_{i - 1}$, where $i \in \left[2,\dots, N\right]$. 

Second, we use a DensePose~\cite{Guler2018DensePose} architecture to estimate correspondences between pixels and parts of the human body, in 3D. 
We apply DensePose~\cite{Guler2018DensePose} to the initial image $P(\mathbf{s})$ and to every frame of a driving video $\left[P(\mathbf{d}_0), \dots, P(\mathbf{d}_N)\right]$, where function $P(\cdot)$ denotes the output of the DensePose. Using this information we obtain a partial correspondence between pixels of any two human images. This correspondence allows us to 
analytically compute the coarse warp grid estimate $W \left[P(\mathbf{s}) \rightarrow P(\mathbf{d}_{i}) \right]$ and  $W \left[P(\mathbf{d}_i) \rightarrow P(\mathbf{d}_{i+1}) \right]$, where $i \in \left[1,\dots,N-1\right]$. The coarse warp grid, in turn, allows us to perform texture transfer and estimate motion flow.
Even though DensePose produces high quality estimates, it is not perfect and sometimes  suffers from artifacts, such as false human detections and missing body parts. Another important drawback of using pose estimators is lack of information regarding clothing. This information is very important to us, since we are trying to map a given person onto a video, while preserving their body shape, facial features, hair and clothing. This motivate us to compute refined warp grid estimates $\overbar{W}\left[P(\mathbf{s}) \rightarrow  P(\mathbf{d}_{i})\right]$ and $\overbar{W}\left[P(\mathbf{d}_i) \rightarrow  P(\mathbf{d}_{i+1})\right]$, where $i \in \left[0,\dots,N-1\right]$ (see Figure~\ref{fig:overview}(b)). We train this component end-to-end using standard image generation losses.

To sum up, our generator $\mathcal{G}$ consists of three blocks: pose encoder $\mathcal{E}$, warp module $\mathcal{W}$ and the decoder $\mathcal{D}$ (see Figure~\ref{fig:overview}(a)). $\mathcal{G}$ generates video frames iteratively one-by-one. First, the encoder $\mathcal{E}$ produces a representation of the driving pose  $\mathcal{E}\left(P(\mathbf{d}_{i+1})\right)$. Then given the source image $\mathbf{s}$, the source pose $P(\mathbf{s})$ and the driving pose $P(\mathbf{d}_{i})$, warp module $\mathcal{W}$ estimates a grid $\overbar{W}\left[P(\mathbf{s}) \rightarrow  P(\mathbf{d}_{i + 1})\right]$, encodes the source image $\mathbf{s}$ and warps this encoded image according to the grid. This gives us a representation $\mathcal{W}\left(\mathbf{s}, P(\mathbf{s}), P(\mathbf{d}_{i + 1}) \right)$. Previous frame is processed in the same way, {\em i.e.}, after transformation we obtain a representation $\mathcal{W}\left(\mathbf{d}_{i}, P(\mathbf{d}_{i}), P(\mathbf{d}_{i + 1}) \right)$. These two representations together with $\mathcal{E}\left(P(\mathbf{d}_{i+1})\right)$ are concatenated and later processed by the decoder $\mathcal{D}$ to produce an output frame $\hat{\mathbf{t}}_{i+1}$.


\subsection{Warp module}
\label{sec:warp-corr}

\textbf{Coarse warp grid estimate}. As described in Section~\ref{sec:over} we use DensePose~\cite{Guler2018DensePose} for coarse estimation of warp grids. For simplicity, we describe only how to obtain warp grid estimates $W\left[P(\mathbf{s}) \rightarrow  P(\mathbf{d}_{i + 1})\right]$; the procedure for $W\left[P(\mathbf{d}_i) \rightarrow  P(\mathbf{d}_{i+1})\right]$ is similar. For each body part of SMPL model~\cite{SMPL:2015} DensePose~\cite{Guler2018DensePose} estimates UV coordinates. However, the UV pixel coordinates in the source image $\mathbf{s}$ may not exactly match with the UV pixel coordinates in the driving frame $\mathbf{d}_{i + 1}$, so in order to obtain a correspondence we use nearest neighbour interpolation. In more detail, for each pixel in the driving frame we find nearest neighbour in the UV space of source image that belongs to the same body part. In order to perform efficient nearest neighbour search we make use of the KD-Trees~\cite{bentley1975multidimensional}.

\vspace{0.1in}
\noindent \textbf{Refined warp grid estimate}. While the coarse warp grid estimation preserves general human body movement, it contains a lot of errors because of self occlusions and imprecise DensePose~\cite{Guler2018DensePose} estimates. Moreover the movement of the person outfit is not modeled. This motivates us to add an additional correction branch $\mathcal{R}$. The goal of this branch is, given source image $\mathbf{s}$, coarse warp grid $W\left[P(\mathbf{s}) \rightarrow  P(\mathbf{d}_{i + 1})\right]$ and target pose $P(\mathbf{d}_{i + 1})$, to predict refined warp grid $ \overbar{W}\left[P(\mathbf{s}) \rightarrow  P(\mathbf{d}_{i + 1})\right]$. This refinement branch is trained end-to-end with the entire framework. Differentiable warping is implemented using bilinear kernel~\cite{jaderberg2015spatial}. The main problem of the bilinear kernel is limited gradient flow, \eg, from each spatial location gradients are propagated only to 4 nearby spatial locations. This makes the module highly vulnerable to the local minimums. One way to address the local minimum problem is good initialization. Having this in mind, we adopt residual architecture for our module, {\em i.e.}, the refinement branch predicts only the correction $\mathcal{R}\left(\mathbf{s}, W\left[P(\mathbf{s}) \rightarrow  P(\mathbf{d}_{i + 1})\right], P(\mathbf{d}_{i + 1})\right)$ which is latter added to the coarse warp grid:
\begin{equation}
    \overbar{W}\left[P(\mathbf{s}) \rightarrow  P(\mathbf{d}_{i + 1})\right]=  W\left[P(\mathbf{s}) \rightarrow  P(\mathbf{d}_{i + 1})\right] +
    \mathcal{R}\left(\mathbf{s}, W\left[P(\mathbf{s}) \rightarrow  P(\mathbf{d}_{i + 1})\right], P(\mathbf{d}_{i + 1})\right).
\end{equation}
Note that since we transform intermediate representations of source image $\mathbf{s}$, the spatial size of the warp grid should be the equal to the spatial size of the representation. In our case the spatial size of the representation is $64 \times 64$. Because of this, and to save computational resources,  $\mathcal{R}$ predicts corrections of size $64 \times 64$; moreover, coarse warp grid $W\left[P(\mathbf{s}) \rightarrow  P(\mathbf{d}_{i + 1})\right]$ is downsampled to the size of $64 \times 64$. Also, since convolutional layers are translation equivariant they can not efficiently process absolute coordinates of coarse grid warp $W\left[P(\mathbf{s}) \rightarrow  P(\mathbf{d}_{i + 1})\right]$. In order to alleviate this issue we input to the network relative shifts, {\em i.e.}, $W\left[P(\mathbf{s}) \rightarrow  P(\mathbf{d}_{i + 1})\right] - I$, where $I$ is an identity warp grid.

\subsection{Training}
\label{sec:train}

Our training procedure is similar to \cite{balakrishnansynthesizing}, however, specifically adopted to take markovian assumption into account. At the training time we sample four frames from the same training video (three of which are consecutive), the indices of these frames are $i, j, j + 1, j + 2$, where $i \in [1,\dots,N]$,  $j \in [1,\dots,N-2]$ and $N$ is the total number of frames in the video. Experimentally we observe that using four frames is the best in terms of temporal consistency and computational efficiency. We treat frame $i$ as the source image $\mathbf{s}$, while the rest are treated as both the driving $\mathbf{d}_i, \mathbf{d}_{i + 1}, \mathbf{d}_{i + 2}$ and the ground truth target $\mathbf{t}_i, \mathbf{t}_{i + 1}, \mathbf{t}_{i + 2}$ frames. We generate the three frames as follows: 
\begin{equation}
 \label{eq:i}
 \mathbf{\hat{t}}_i = \mathcal{G}\left(\mathbf{s}, P(\mathbf{s}), \mathbf{s}, P(\mathbf{s}), P(\mathbf{d}_{i})\right),   
\end{equation}
for the first frame, where the source frame is treated as the "previous" frame, and for the rest: 
\begin{eqnarray}
    \label{eq:i-plus-1}
    \mathbf{\hat{t}}_{i + 1}  & = & \mathcal{G}\left(\mathbf{s}, P(\mathbf{s}), \mathbf{\hat{t}}_{i}, P(\mathbf{d}_{i}), P(\mathbf{d}_{i + 1})\right), \nonumber \\
    \mathbf{\hat{t}}_{i + 2} & = & \mathcal{G}\left(\mathbf{s}, P(\mathbf{s}), \mathbf{\hat{t}}_{i+1}, P(\mathbf{d}_{i+1}), P(\mathbf{d}_{i + 2})\right).
\end{eqnarray}
This formulation has low memory consumption, but at the same time allows standard pose-guided image generation which is needed to produce the first target output frame. Note that if in Eq.~\ref{eq:i-plus-1} we use real previous frame $\mathbf{d}_i$ the generator will ignore the source image $\mathbf{s}$, since $\mathbf{d}_i$ is much more similar to $\mathbf{d}_{i + 1}$ than $\mathbf{s}$.

\vspace{0.1in}
\noindent \textbf{Losses}. We use a combination of losses from pix2pixHD~\cite{pix2pixHD} model. We employ the least-square GAN \cite{mao2017least} for the adversarial loss:
\begin{eqnarray}
  \mathcal{L}^\mathcal{D}_\mathrm{gan} (\mathbf{t}_i, \mathbf{\hat{t}}_i) & = & (\mathcal{C}(\mathbf{t}_i) -1)^2 + \mathcal{C}(\mathbf{\hat{t}}_i)^2, \\
\mathcal{L}^\mathcal{G}_\mathrm{gan} (\mathbf{\hat{t}}_i) & = & (\mathcal{C}(\mathbf{\hat{t}}_i)-1)^2,
\label{eq:GAN-loss}
\end{eqnarray}
where $\mathcal{C}$ is the patch bases critique~\cite{pix2pixHD, pix2pix2016},

To drive image reconstruction we also employ a feature matching~\cite{pix2pixHD} and perceptual~\cite{johnson2016perceptual} losses:
\begin{equation}
\mathcal{L}_\mathrm{rec} (\mathbf{t}_i, \mathbf{\hat{t}}_i) =\Vert \mathcal{N}_k(\mathbf{\hat{t}}_i)-\mathcal{N}_k(\mathbf{t}_i)\Vert_1, 
\end{equation}
where $\mathcal{N}_k$ is feature representation from $k$-th layer of the network, for perceptual loss this is VGG-19~\cite{DBLP:journals/corr/SimonyanZ14a} and for the feature matching it is the critique $\mathcal{C}$. The total loss is given by:

\begin{equation}
\mathcal{L} = \sum_i\mathcal{L}^\mathcal{G}_\mathrm{gan} (\mathbf{\hat{t}}_i) + \lambda \mathcal{L}_\mathrm{rec} (\mathbf{t}_i, \mathbf{\hat{t}}_i),
\end{equation}
following~\cite{pix2pixHD} $\lambda=10$.


\section{Experiments}
\label{sec:experiments}
We have conducted an extensive set of experiments to evaluate the proposed DwNet. We first describe our newly collected dataset, then we compare our method with previous state-of-the-art models for pose-guided human video synthesis and for pose-guided image generation. We show superiority of our model in aspects of realism and temporal coherency. Finally, we evaluate the contributions of each architecture choice we made and show that each part of the model positively contributes to the results.

\subsection{The Fashion Dataset}
We introduce a new Fashion dataset containing 500 training and 100 test videos, each containing roughly 350 frames. Videos from our dataset are of a single human subject and characterized by the high resolution and static camera. 
Most importantly, clothing and textures are diverse and cover large space of possible appearances. The dataset is publicly released at: \url{https://vision.cs.ubc.ca/datasets/fashion/}. 

\subsection{Setup}
\label{sec:setup}
\subsubsection{Datasets} We conduct our experiments on the proposed Fashion and Tai-Chi \cite{tulyakov2017mocogan} datasets. The latter is composed of more than 3000 tai-chi video clips downloaded from YouTube. In all previous works~\cite{siarohin2019animating, tulyakov2017mocogan}, the smaller $64 \times 64$ pixel resolution version of this dataset has been used; however, for our work we use $256 \times 256$ pixel resolution. The length varies from 128 to 1024 frames per video. Number of videos per train and test sets are 3049 and 285 respectively. 

\subsubsection{Evaluation metrics}
There is no consensus in the community on a single criterion for measuring quality of the generated videos from the perspective of realism, texture similarity and temporal coherence. Therefore we choose a set widely accepted evaluation metrics to measure performance.

\vspace{0.1in}
\noindent
\textbf{Perceptual loss.} The texture similarity is measured using a perceptual loss.
Similar to our training procedure, we use VGG-19~\cite{DBLP:journals/corr/SimonyanZ14a} network as a feature extractor and then compute $L_1$ loss between the extracted features from the real and generated frames.

\vspace{0.1in}
\noindent
\textbf{FID.} We use Frecht Inception Distance~\cite{heusel2017gans} (FID) to measure realism of the individual frames. FID is known to be a widely used metric for comparison of the GAN-based methods.

\vspace{0.1in}
\noindent
\textbf{AKD.} We evaluate if the motion is correctly transferred by the means of Average Keypoint Distance (AKD) metric~\cite{siarohin2019animating}. Similarly to~\cite{siarohin2019animating} we employ human pose estimator~\cite{cao2017realtime} and compare average distance between ground truth and detected keypoints. Intuitively this metric measures if the person moves in the same way as in the driving video. We do not report Missing keypoint rate (MKR) because it similar and close to zero for all methods.

\vspace{0.1in}
\noindent
\textbf{User study.} We conduct a user study to measure the overall temporal coherency and quality of the synthesised videos. For the user study we exploit Amazon Mechanical Turk (AMT). On AMT we show users two videos (one produced by DwNet and another by a competing method) in random order and ask users to choose one, which has higher realism and consistency. To conduct this study we follow the protocol introduced in Siarohin ~\etal~\cite{siarohin2019animating}.

\subsection{Implementation details}
All of our models are trained for two epochs. In our case epoch denotes a full pass through the whole set of video frames, where each sample from the dataset is a set of four frames, as explained in the Section~\ref{sec:train}. 
We train our model starting with the learning rate 0.0002 and bring it down to zero during the training procedure. Generally, our model is similar to Johnson~\etal~\cite{johnson2016perceptual}. Novelties of the architecture such as pose encoder $\mathcal{E}$ and the appearance encoder $\mathcal{A}$ both contain 2 downsampling Conv layers. Warp module's refine branch $\mathcal{R}$ is also based on 2 Conv layers and additional 2 ResNet blocks. Our decoder $\mathcal{D}$ architecture is made out of 9 ResNet blocks and 2 upsampling Conv layers. We perform all our training procedures on a single GPU (Nvidia GeForce GTX 1080). Our code will be released. 

\begin{figure}[t]
\centering
\includegraphics[width=1.0\textwidth]{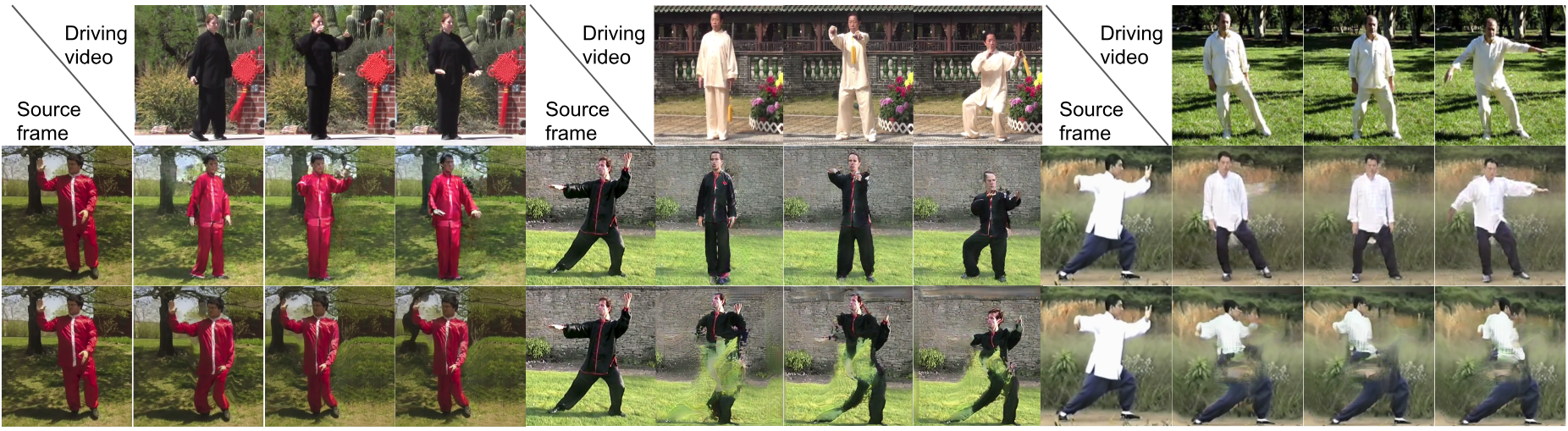}
\caption{{\bf Qualitative Results on Tai-Chi Dataset.} First row illustrates the driving videos; second row are results of our method; third row are results obtained with  Monkey-Net~\cite{siarohin2019animating}. }
\label{fig:equi}
\end{figure}

\subsection{Comparison with the state-of-the-art}
\label{sec:sota}

We compare our framework with the current state-of-the-art method for motion transfer MonkeyNet~\cite{siarohin2019animating}, which solves a similar problem for the human synthesis. The first main advantage of our method, compared to MonkeyNet, is ability to generate frames with a higher resolution. Originally, MonkeyNet was trained on $64 \times 64$ size frames. However, to conduct fair experiments we re-train MonkeyNet from scratch to produce the same size images with our method. Our task is quite novel and there is limited number of baselines. To this end, we also compare with Coordinate Inpainting framework~\cite{grigorev2019coordinate} which is state-of-the-art for image (not video) synthesis, {\em i.e.}, synthesise of a new image of a person based on a single image. Even though this framework solves a slightly different problem, we still choose to compare with it, since it is similarly leverages DensePose~\cite{Guler2018DensePose}. This approach doesn't have any explicit background handling mechanisms therefore there is no experimental results on a Tai-Chi dataset. Note that since authors of the paper haven't released the code for the method we were only able to run our experiments on a pre-trained network.

\begin{figure}[h]
    \centering
    \includegraphics[width=1.0\textwidth]{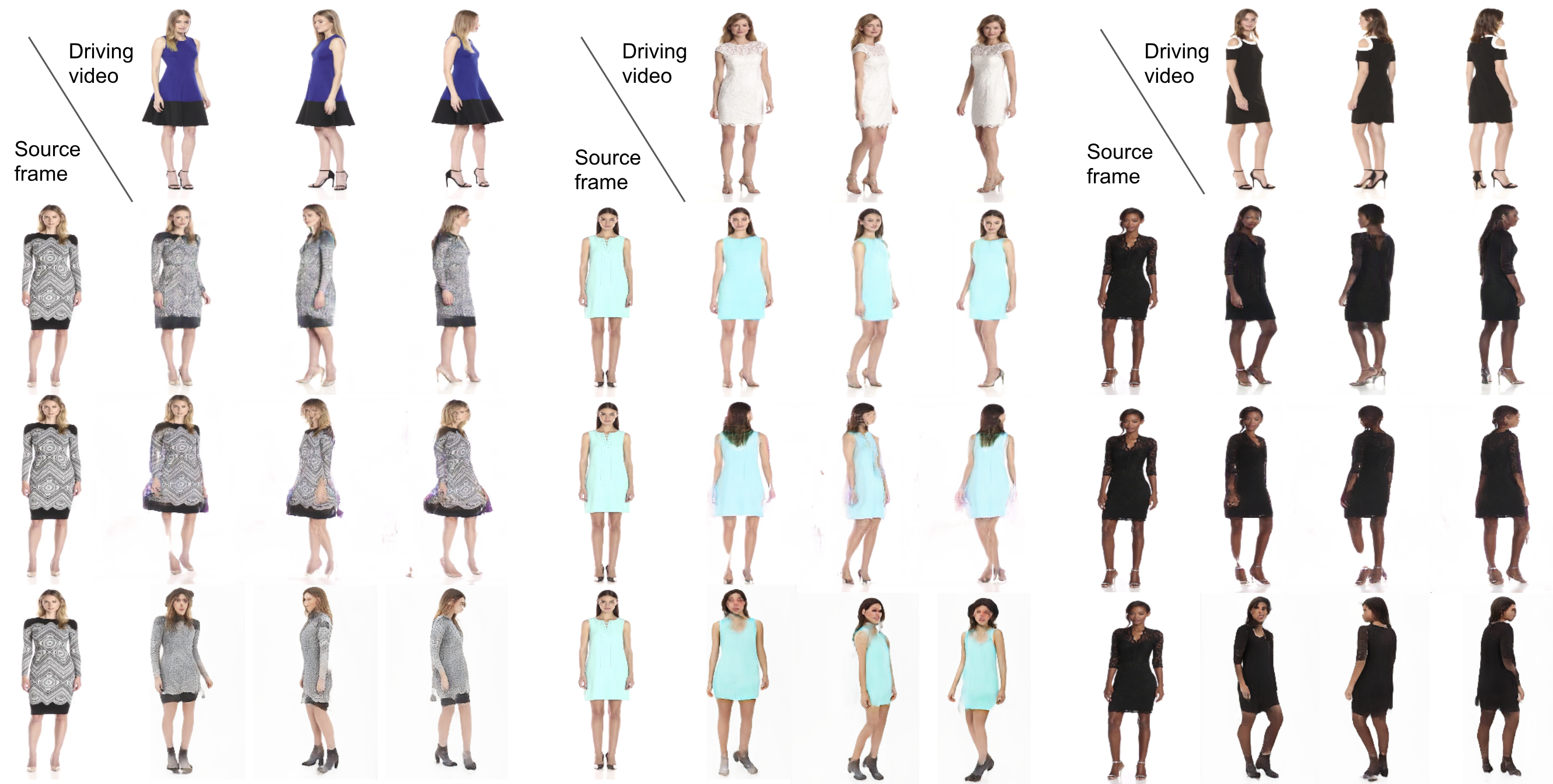}
    \caption{{\bf Qualitative Results on Fashion Dataset.} 
    First row illustrates the driving videos; second row are results of our method; third row are results obtained with Monkey-Net~\cite{siarohin2019animating}; 
    forth row are results of Coordinate Inpainting~\cite{grigorev2019coordinate}. Please zoom in for detail.}
\label{fig:Equilibria}
\end{figure}

The quantitative comparison is reported in Table~\ref{tab:recSota}. Our approach outperforms  MonkeyNet and Coordinate Inpainting on both datasets and according to all metrics. With respect to MonkeyNet, this can be explained by its inability to grasp complex human poses, hence it completely fails on the Tai-Chi dataset which contains large variety of non-trivial motions. This can be further observed in Figure~\ref{fig:equi}. MonkeyNet simply remaps person from the source image without modifying the pose. In Figure~\ref{fig:Equilibria} we can still observe a large difference in terms of the human motion consistency and realism. Unlike our model, MonkeyNet produces images with missing body parts. For Coordinate Inpainting, poor performance could be explained by the lack of temporal consistency, since (unlike our method) it generates each frame independently and hence lacks consistency in clothing texture and appearance. Coordinate Inpainting is heavily based on the output of the DensePose and doesn't correct resulting artifacts, like is done in our model using refined warp grid estimate. As one can see from Figure~\ref{fig:Equilibria} the resulting frames are blurry and inconsistent in small details. This can also explains why such a small percentage of users prefer results of Coordinate Inpainting. The user study comparison is reported in Table~\ref{tab:user_study} where we can observe that videos produced by our model were significantly more often preferred by users in comparison to the videos from the competitors models. 

\begin{table}[t]
\centering
\resizebox{\textwidth}{!}{
    \begin{tabular}{c|ccc|ccc}
    \toprule
        & \multicolumn{3}{c|}{Fashion} & \multicolumn{3}{c}{Tai-Chi} \\
        & Perceptual ($\downarrow$)& FID ($\downarrow$) & AKD ($\downarrow$) & Perceptual ($\downarrow$) & FID ($\downarrow$) & AKD ($\downarrow$) \\
        \midrule
        Monkey-Net [3] & 0.3726  & 19.74 & 2.47 & 0.6432 & 94.97 & 10.4 \\
        Coordinate Inpainting [4] & 0.6434 & 66.50 & 4.20 & - & - & - \\
        Ours & \bf 0.2811 & \bf 13.09 &  \bf 1.36  & \bf 0.5960 & \bf 75.44 &  \bf 3.77 \\
        \bottomrule
    \end{tabular}
}
    \caption{{\bf Quantitative Comparison with the State-of-the-Art.} Performance on Fashion and Tai-chi datasets is reported in terms of the Perceptual Loss, Frechet Inception Distance (FID) and Average Keypoint Distance (AKD)}
    \label{tab:recSota}
\end{table}

\begin{table}[t]
\centering
\resizebox{0.7\textwidth}{!}{
    \begin{tabular}{c|c|c}
    \toprule
        & \multicolumn{1}{c|}{Fashion} & \multicolumn{1}{c}{Tai-Chi} \\
        & User-Preference ($\uparrow$) & User-Preference ($\uparrow$) \\
        \midrule
        Monkey-Net [3] & 60.40\% & 85.20\% \\
        Coordinate Inpainting [4] & 99.60\% &  -\\
        \bottomrule
    \end{tabular}
}
\caption{{\bf User study.} Percentage of the time our method is preferred over one of the competing approaches.}
\label{tab:user_study}

\end{table}

\subsection{Ablation}
\label{sec:abl}

\begin{figure}[h]
\begin{minipage}{0.5\textwidth}
\includegraphics[width=\textwidth]{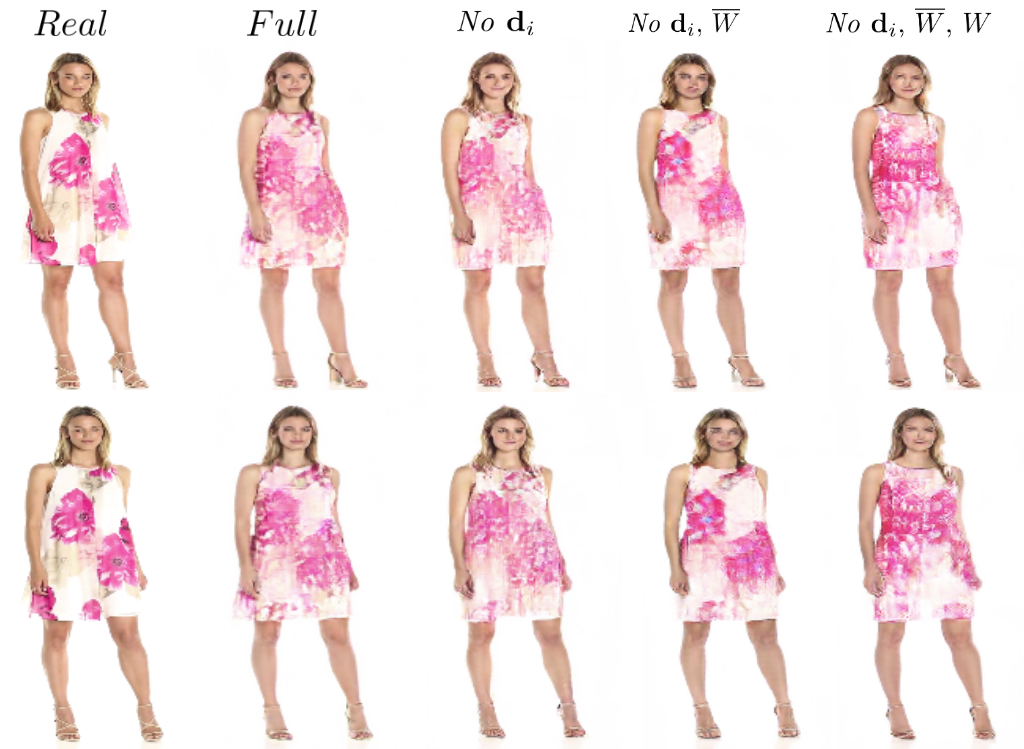}
\end{minipage}
\begin{minipage}{0.3\textwidth}
    \scriptsize
    \begin{tabular}{c|cc}
    \toprule
        & \multicolumn{2}{c}{Fashion} \\
        & Perceptual ($\downarrow$) & FID ($\downarrow$)\\
        \midrule
         $\text{\textit{No} }\mathbf{d}_i\text{, }\overline{W}\text{, }W$ & 0.29 & 17.18  \\
         $\text{\textit{No} }\mathbf{d}_i\text{, }\overline{W}$ & 0.29 & 15.37 \\
        $\text{\textit{No} }\mathbf{d}_i $ & 0.29 & 15.05 \\
        \text{\textit{Full}} & \bf 0.28 & \bf 13.09 \\
        \bottomrule
    \end{tabular}
\end{minipage}
\caption{{\bf Ablations on Fashion Dataset.} On the (left) are qualitative result from the ablated methods; on the (right) are corresponding quantitative evaluations. See text for details.}
\label{fig:ablations}
\end{figure}

Table in Figure~\ref{fig:ablations} (right) shows the contribution of each of the major architecture choices, {\em i.e.}, markovian assumption ($\text{\textit{No} }\mathbf{d}_i$), refined warp grid estimate ($\text{\textit{No} }\mathbf{d}_i\text{, }\overline{W}$) and coarse warp grid estimate ($\text{\textit{No} }\mathbf{d}_i\text{, }\overline{W}\text{, }W$). For these experiments we remove mentioned parts from DwNet and train the resulting model architectures. As expected, removing markovian assumption, {\em i.e.}, not conditioning on the previous frame, leads to a worse realism and lower similarity with the features of a real image. Mainly it is because this leads to a loss of temporal coherence. Further removal of both  warping grid estimators, in the generation pipeline, results in worse performance in the FID score. Perceptual loss is not affected by this change which can be explained by the fact that warp mostly results in removal of the artifacts and naturalness of the texture on a person. In Figure~\ref{fig:ablations} (left) we see the qualitative reflection of our quantitative results. Full model produces the best results, third column shows misalignment between textures of two frames. The architecture without a refined warp produces less realistic results, with a distorted face. Lastly, an architecture without any warp produces blurry, unrealistic results with an inconsistent texture.

%
%

\section{Conclusion}
\label{sec:conclusion}

In this paper we present DwNet a generative architecture for pose-guided video generation. Our model can produce high quality videos, based on the source image depicting human appearance and the driving video with another person moving. We propose novel markovian modeling to address temporal inconsistency, that typically arises in video generation frameworks. Moreover we suggest novel warp module that is able to correct warping errors. We validate our method on two video datasets, and we show superiority of our method over the baselines. Some possible future directions may include multiple source generation and exploiting our warp correction for improving DensePose~\cite{Guler2018DensePose} estimation.

\bibliography{egbib}
\end{document}